 \documentclass[pmlr,twocolumn,10pt]{jmlr} % W&CP article

\usepackage{booktabs}
\usepackage{siunitx}
\usepackage{multirow}
\usepackage[normalem]{ulem}
\useunder{\uline}{\ul}{}
\usepackage{rotating}

\usepackage[switch]{lineno}

\newcommand{\equal}[1]{{\hypersetup{linkcolor=black}\thanks{#1}}}

\theorembodyfont{\upshape}
\theoremheaderfont{\scshape}
\theorempostheader{:}
\theoremsep{\newline}

\jmlrvolume{297}
\jmlryear{2025}
\jmlrworkshop{Machine Learning for Health (ML4H) 2025} % W&CP title

 \title[Evaluating MLLMs for Pathology Localization]{Beyond Diagnosis: Evaluating Multimodal LLMs for Pathology Localization in Chest Radiographs}

\author{%
\Name{Advait Gosai}\equal{These authors contributed equally}\thanks{Work done while intern at Dana-Farber Cancer Institute} \Email{advait\_gosai@berkeley.edu}\\
\addr University of California, Berkeley
\AND
\Name{Arun Kavishwar}\footnotemark[1] \Email{arun\_kavishwar@alumni.brown.edu}\\
\addr Dana-Farber Cancer Institute
\AND
\Name{Stephanie L. McNamara} \Email{slmcnamara@mgh.harvard.edu}\\
\addr Massachusetts General Hospital
\AND
\Name{Soujanya Samineni} \Email{soujanya\_samineni@dfci.harvard.edu}\\
\addr Dana-Farber Cancer Institute
\AND
\Name{Renato Umeton} \Email{renato.umeton@stjude.org}\\
\addr St. Jude Children's Research Hospital
\AND
\Name{Alexander Chowdhury} \Email{alexander\_chowdhury@dfci.harvard.edu}\\
\addr Dana-Farber Cancer Institute
\AND
\Name{William Lotter} \Email{lotterb@ds.dfci.harvard.edu}\\
\addr Dana-Farber Cancer Institute, Brigham and Women's Hospital, \& Harvard Medical School
}

\begin{document}

\maketitle

\begin{abstract}
Recent work has shown promising performance of frontier large language models (LLMs) and their multimodal counterparts in medical quizzes and diagnostic tasks, highlighting their potential for broad clinical utility given their accessible, general-purpose nature. However, beyond diagnosis, a fundamental aspect of medical image interpretation is the ability to localize pathological findings. Evaluating localization not only has clinical and educational relevance but also provides insight into a model’s spatial understanding of anatomy and disease.
Here, we systematically assess two general-purpose MLLMs (GPT-4 and GPT-5) and a domain-specific model (MedGemma) in their ability to localize pathologies on chest radiographs, using a prompting pipeline that overlays a spatial grid and elicits coordinate-based predictions. 
Averaged across nine pathologies in the CheXlocalize dataset, GPT-5 exhibited a localization accuracy of 49.7\%, followed by GPT-4 (39.1\%) and MedGemma (17.7\%), all lower than a task-specific CNN baseline (59.9\%) and a radiologist benchmark (80.1\%). Despite modest performance, error analysis revealed that GPT-5's predictions were largely in anatomically plausible regions, just not always precisely localized. 
GPT-4 performed well on pathologies with fixed anatomical locations, but struggled with spatially variable findings and exhibited anatomically implausible predictions more frequently. MedGemma demonstrated the lowest performance on all pathologies, but showed improvements when provided examples through few shot prompting.  
Our findings highlight both the promise and limitations of current MLLMs in medical imaging and underscore the importance of integrating them with task-specific tools for reliable use.
\end{abstract}


\paragraph*{Data and Code Availability}
This study used the public CheXlocalize dataset~\citep{Saporta2022-lo}. Code is available at \url{https://github.com/lotterlab/mllm_localization}.

\paragraph*{Institutional Review Board (IRB)}
This research does not require IRB approval. 

\section{Introduction}

As AI advances from task-specific to generalist models, there is growing interest in evaluating frontier models in medicine, including in medical imaging with the rise of multimodal large language models (MLLMs). 
Recent studies assessing GPT-4, a pioneering MLLM, on imaging-based quizzes and diagnostic tasks have yielded mixed results~\citep{Hayden2024-sc, Brin2024-gp, Strotzer2024-wo, Zhou2024-qq, Suh2024-qw, Eriksen2023-ye}. 
For instance, \citet{Suh2024-qw} reported that GPT-4 achieved clinical-level performance on \textit{NEJM Image Challenges}, a multiple-choice, VQA dataset; however, \cite{Jin2024-ai} found that the model's rationale for its choices were often flawed. 
Given the clinical potential of accessible, general-purpose MLLMs, these findings highlight the need to evaluate their performance across diverse tasks and to move beyond diagnostic accuracy toward a deeper understanding of their reasoning. Furthermore, it remains unclear if domain-specific MLLMs are required to address the challenges of general purpose models, or whether these challenges can instead be overcome with newer model versions.

%Furthermore, it remains uncertain whether domain-specific MLLMs are needed to overcome the limitations of general-purpose models, or if such challenges can instead be addressed through newer model iterations.

A fundamental aspect of medical image interpretation is not only providing a diagnostic impression but also localizing findings associated with a diagnosis. 
In computer vision, diagnosis is typically framed as a classification task, whereas localization is approached through segmentation or bounding-box generation (i.e., detection) over relevant regions. 
In medical AI systems, these two tasks are treated as distinct regulatory categories: computer-aided diagnosis (CADx) for classification and computer-aided detection (CADe) for localization~\citep{McNamara2024-oc}. 
Critically, an AI model may perform well at classification, but struggle with precisely localizing the associated findings, or conversely struggle with distinguishing whether a finding is benign or pathological, but excel at identifying its location.
While recent studies have evaluated the diagnostic performance of MLLMs, their ability to localize pathological findings (e.g., CADe) remains largely unexplored. 
Beyond potential clinical applications, evaluating MLLMs in localization can provide insights into their underlying clinical intuition and medical knowledge. 

\begin{figure*}[!h]
\centering 
\includegraphics[width=.85\textwidth]{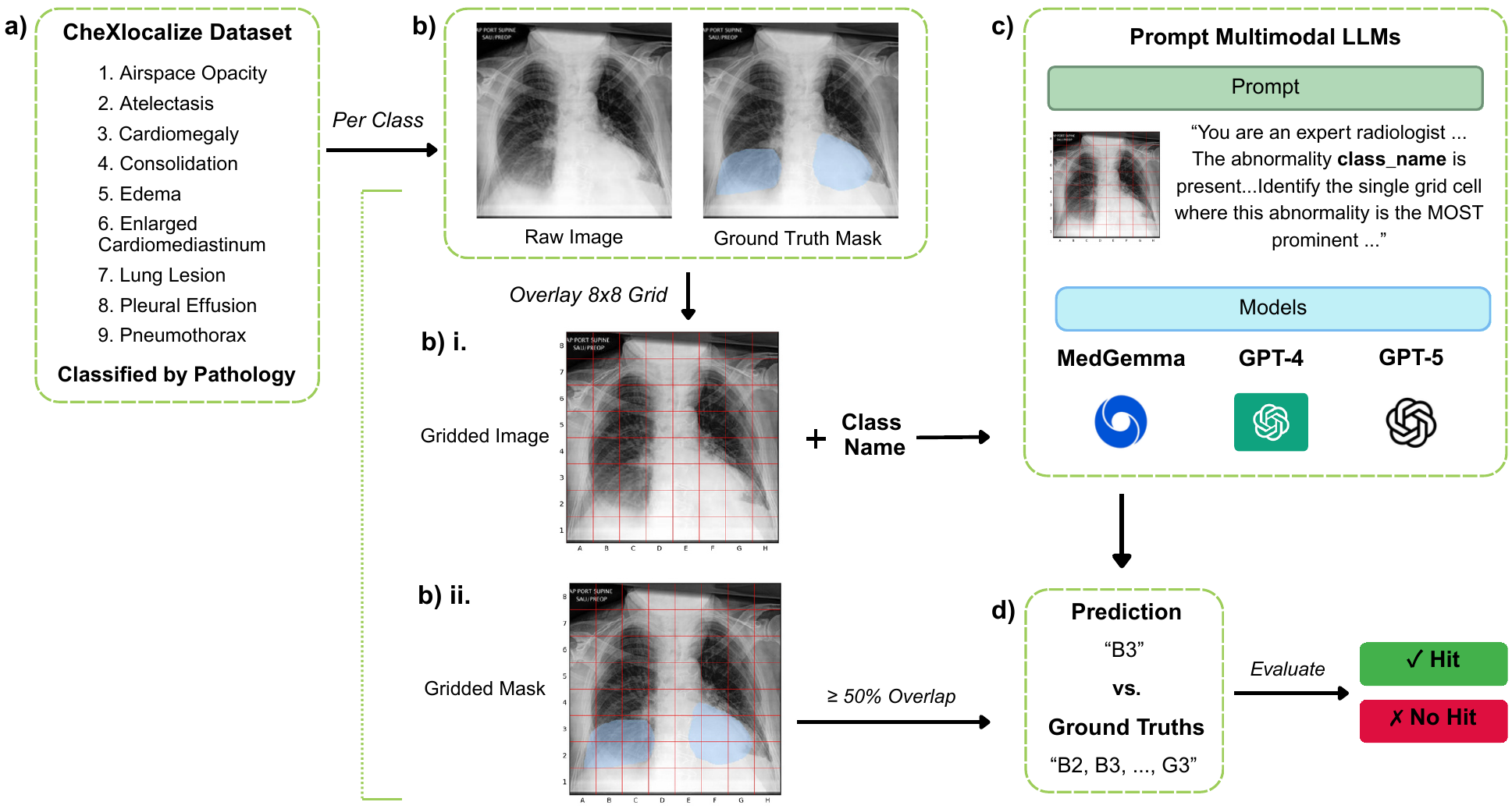} 
\vspace{-8pt}
\caption{Study Overview. a) The nine pathologies in the CheXlocalize dataset were utilized. b) For each pathology class, a gridded image was produced for each radiograph where the pathology was present. c) Three multimodal LLMs were prompted to identify the grid cell where the stated pathology is most prominent. d) The predicted grid cells were compared to the ground truth cells, defined as cells with $\ge$50\% overlap with the ground truth mask.}
\label{fig:methods_fig} 
\end{figure*}

In this study, we systematically evaluate the ability of three MLLMs to localize pathologies on chest radiographs: two generalist models (GPT-4 \citep{openai2024gpt4technicalreport} and GPT-5 \citep{gpt5}) and a domain-specific model (MedGemma \citep{medgemma}). 
To do so, we develop a prompting strategy that overlays a grid on the image and asks the model to predict coordinates corresponding to the pathology's location. 
We assess performance using the CheXlocalize dataset~\citep{Saporta2022-lo}, benchmarking against radiologists and a task-specific CNN model. 
To further explore the models' medical understanding, we analyze the distribution of predicted locations for each pathology and categorize mislocalizations based on their clinical plausibility. 
By benchmarking from a new angle--disease localization--and performing detailed error analysis, we sought to elucidate the spatial understanding of anatomy and disease of frontier MLLMs, probing both the potential for clinical utility and current limitations. 
%By moving beyond classification tasks, our analysis provides insights into the spatial reasoning abilities of MLLMs in medical imaging, highlighting both strengths and limitations.

\section{Related Work}

Several recent studies have assessed general-purpose MLLMs such as GPT-4 in medical image interpretation~\citep{Jin2024-ai, Hayden2024-sc, Brin2024-gp, Strotzer2024-wo, Zhou2024-qq, Suh2024-qw, Eriksen2023-ye}. 
On 190 cases from the \textit{Diagnosis Please} challenge, \cite{Suh2024-qw} found that GPT-4 achieved diagnostic performance comparable to radiologists. 
GPT-4 also demonstrated strong accuracy on the \textit{NEJM} Image Challenges, a multiple-choice diagnostic quiz for medical professionals~\citep{Jin2024-ai, Eriksen2023-ye}. 
However, \cite{Jin2024-ai} reported that while GPT-4 performed well on these tasks, its rationales were often flawed when explicitly prompted to explain its reasoning. 
On multiple-choice questions from the \textit{American College of Radiology Diagnostic Radiology In-Training Examination}, GPT-4 achieved high accuracy on text-only questions (81.5\%) but struggled with image-based ones (47.8\%). 
Its diagnostic accuracy on real-world radiologic imaging cohorts has also shown considerable variability \citep{Strotzer2024-wo, Brin2024-gp, Zhou2024-qq}.

\section{Methods}

While prior research has focused primarily on the diagnostic capabilities of MLLMs, our study examines spatial localization accuracy for pathologies in medical images that the models are explicitly prompted to recognize. We utilize chest radiographs from CheXlocalize~\citep{Saporta2022-lo}, a subset of the CheXpert~\citep{chexpert} dataset annotated with pixel-level masks for each of its classes.
We focus on this task because it serves as both a common benchmark for medical AI and a clinically important application. Figure~\ref{fig:methods_fig} contains an overview of the approach.

\subsection{CheXlocalize Dataset}
The CheXlocalize dataset includes chest radiograph images collected from Stanford Hospital in both inpatient and outpatient settings~\citep{chexpert}. 
The dataset contains expert radiologist annotations for the localization of 10 distinct findings~\citep{Saporta2022-lo}. These ground-truth annotations come in the form of pixel-wise segmentation maps, enabling precise benchmarking. 
The validation and test splits of this dataset consist of 234 radiographs from 200 patients and 668 radiographs from 500 patients, respectively.
All results reported in the current study are based on the test split.
The image counts by pathology and view position are contained in the Appendix. 

The 10 classes annotated in CheXlocalize are Atelectasis, Cardiomegaly, Consolidation, Edema, Enlarged Cardiomediastinum, Lung Lesion, Lung Opacity, Pleural Effusion, Pneumothorax, and Support Devices. In this work, we include the first 9 classes, excluding Support Devices to focus exclusively on pathology localization. Our evaluation encompasses both frontal (anterior-posterior and posterior-anterior) and lateral view radiographs available in this dataset for a comprehensive assessment. We compare our results on the test split to an expert human benchmark and a convolutional neural network (CNN) baseline reported alongside the dataset in \cite{Saporta2022-lo}.
Both the ground truth annotations and the human benchmarking assessment were performed by radiologists, where the benchmarking effectively quantifies a performance ceiling of inter-reader variability. Specifically, ground truth annotations were performed by two board-certified radiologists. Three separate radiologists then also performed annotations, which were compared to the ground truth annotations to create the human performance benchmark.
The CNN baseline consists of a DenseNet121~\citep{densenet} classification model trained on CheXpert. Localization predictions were obtained through a GradCAM-based~\citep{gradcam} saliency pipeline.
As detailed below, we compare the MLLMs to the `hit rate' performance metric for the human and CNN benchmarks, which was quantified by \cite{Saporta2022-lo} as whether the most representative point identified by the CNN saliency method or human benchmark annotations fell within the ground truth segmentation. 

\subsection{Prompting Technique}
To evaluate MLLM localization capabilities, we designed a structured prompting pipeline by which each chest radiograph was digitally overlaid with a standardized grid. The grid divided each image into equally-sized cells, labeled alphanumerically to facilitate text-based spatial referencing. To do so, each radiograph image was first centrally cropped and resized to $256\times256$ pixels, a common pre-processing strategy for computer vision models. The MLLM was then tasked with identifying the specific grid cell that was most representative of the given pathology.
For the core results, we use an 8x8 grid overlay ($32\times32$ pixels per cell), which was chosen to balance the resolution size with the typical size of the pathology masks.
Analysis using a higher resolution grid (16x16 cells) is also presented in the Appendix for sensitivity analysis.

We used the same prompt for all MLLMs, which explicitly stated that the given pathology was present in the image in order to make localization, not identification, the sole objective (see Appendix for full prompt). Because a single radiograph could exhibit multiple pathologies, we generated separate copies of each image for every present pathology and performed individual queries for each pathology-image pair. This approach enabled independent evaluation of localization performance for each pathology. 

\begin{figure*}[t]
\centering 
\includegraphics[width=.9\textwidth]{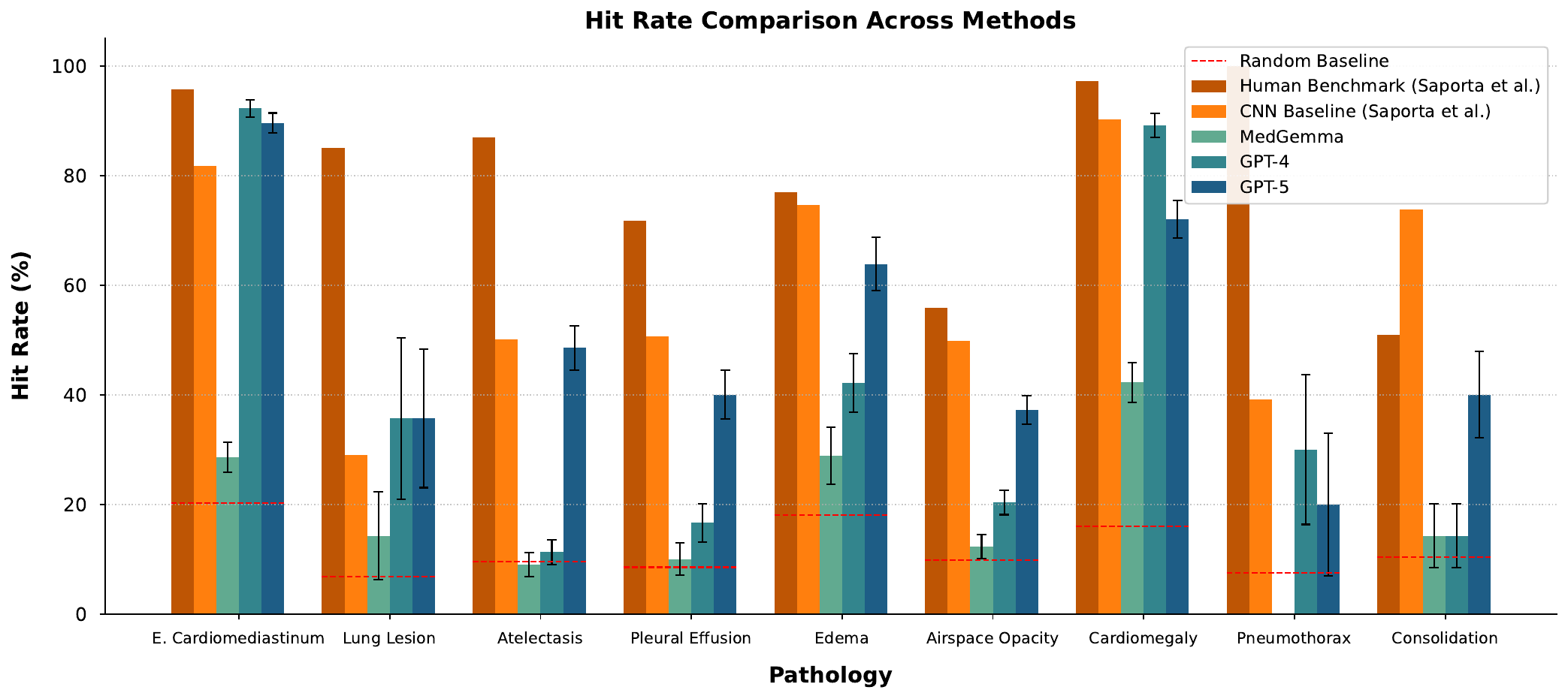} 
\vspace{-12pt}
\caption{Hit rate by pathology. Pathologies are ordered by the relative performance difference between GPT-5 and the CNN baseline. Error bars represent standard deviation via bootstrapping. }
\label{fig:hit_rate_summary} 
\end{figure*}

%The performances of GPT-4, GPT-5, and MedGemma are compared to the clinician benchmark and CNN baseline reported by \cite{Saporta2022-lo}.

\subsection{Evaluation Metrics}
Performance evaluation was based on a `hit rate' criterion, designed to mirror the pointing game methodology introduced by \cite{Saporta2022-lo}. A prediction (represented by a single grid cell returned by the MLLM) was considered a `hit' if at least 50\% of the grid cell overlapped with the ground truth segmentation provided in the CheXlocalize annotations for the queried pathology. Predictions falling below this overlap threshold were classified as misses. In the rare scenario where the ground truth mask contained no cells with $\ge$50\% overlap (4.4\% of all images), the prediction was deemed a hit if the cell contained any amount of ground truth mask. This could occur, for instance, if the segmentation mask was very small. Our defined hit rate metric enables meaningful comparison with the results reported by \cite{Saporta2022-lo}, wherein the pointing game methodology for their human benchmark awarded credit when the identified representative point fell within the ground-truth segmentation region. We report the average hit rate for each pathology and MLLM, alongside standard deviation error bars using bootstrapping (1,000 samples). 

\subsection{Error Analysis}
Predictions not meeting the hit criteria were qualitatively reviewed by a radiologist and categorized into three distinct error types:
\begin{itemize}
\item \textbf{Partial hit}: Predicted cell partially overlapped ($\le$50\%) with the ground truth annotation.
\item \textbf{Position error}: Predicted cell overlapped with anatomically plausible but incorrect regions for the pathology.
\item \textbf{Anatomy error}: Predicted cell was anatomically implausible for the given pathology, suggesting a potential misunderstanding of chest anatomy and/or the pathology.
\end{itemize}
For consistency, the error analysis was only performed on frontal radiographs, which represent 87\% of images in the test set. If a model produced more than 50 complete misses for a given pathology, only 50 were reviewed, and the proportion of position versus anatomy errors was estimated from this subset.

%\vspace{-15pt}
\subsection{Models}

GPT-4 and GPT-5 were executed using the OpenAI API. The model version for GPT-4 was \verb|gpt-4o-2024-05-13|, configured at a temperature of 0 for reproducibility. The model version for GPT-5 was not obtainable via the API, but it was run within a month of its release on 2025-08-07, and the only relevant configurable parameter available was the reasoning effort, which was kept at the default level of medium. MedGemma was executed locally on 2 H100 GPUs using the Ollama platform. The 27B, instruction-tuned variant was used (accessed via puyangwang/medgemma-27b-it:q8). The temperature for MedGemma was left at a default value of 0.8. A temperature of 0 was also initially considered, but this resulted in almost no variability in the predicted grid cells (96.7\% of predictions in either D4 or C4); thus the default temperature was deemed the most appropriate assessment for MedGemma. 

\section{Results}

%We evaluated the ability of three MLLMs -- GPT-4, GPT-5, and MedGemma -- in localizing pathologies in chest radiographs using the CheXlocalize dataset.
%The models was asked to predict the grid coordinates of a pathology that was stated to be present for 9 different pathologies. 

\subsection{Hit Rate Performance}
Figure~\ref{fig:hit_rate_summary} summarizes the hit rate performance of the MLLMs compared to the results reported by \cite{Saporta2022-lo}, consisting of an expert human (radiologist) benchmark and a convolutional neural network (CNN) baseline. A random baseline is also presented, which reflects the expected performance if cell selections were made uniformly at random over the grid. The random baseline is computed per pathology by first computing the chance level for each image (\# ground truth cells / total cells) and then averaging across all images for that pathology.

Between the three MLLMs, GPT-5 had the highest average hit rate across pathologies (49.7\%), followed by GPT-4 (39.1\%), and MedGemma (17.7\%). All models underperformed the CNN baseline and human benchmark on average, which showed 59.9\% and 80.1\% average hit rates, respectively. The average random baseline across pathologies was 11.9\%.

Different trends were observed for different pathologies. GPT-4 and GPT-5 both outperformed the CNN baseline on enlarged cardiomediastinum (hit rate of 92.3\% for GPT-4, 89.6\% for GPT-5, 81.8\% for CNN). GPT-5 showed the greatest improvements ($\ge$20\% higher hit rates) over GPT-4 on atelectasis, consolidation, edema, and pleural effusion, though it still underperformed the CNN baseline on these pathologies. GPT-4 exhibited its highest performance over GPT-5 for cardiomegaly (hit rate of 89.1\% vs. 72.0\%). MedGemma underperformed for each pathology, with its largest margin over chance occurring for cardiomegaly. 
These overall findings were robust to the resolution of the grid overlay, where halving the size of each cell showed similar patterns (Appendix Figure~\ref{fig:gridsize_performance}).

\subsection{Error Categorization}
% To gain insights into the MLLMs' performance and general medical understanding, each miss for the three MLLMs was qualitatively reviewed by a radiologist and categorized into three distinct error types:
% \begin{itemize}
% \item \textbf{Partial hit}: Predicted cell partially overlapped ($\le$50\%) with the ground truth annotation.
% \item \textbf{Position error}: Predicted cell overlapped with anatomically plausible but incorrect regions for the pathology.
% \item \textbf{Anatomy error}: Predicted cell  was anatomically implausible for the given pathology, suggesting a potential misunderstanding of chest anatomy and/or the pathology.
% \end{itemize}
% For consistency, the error analysis was only performed on frontal radiographs, which represent 87\% of images in the test set. If a model produced more than 50 complete misses for a given pathology, only 50 were reviewed, and the proportion of position versus anatomy errors was estimated from this subset.

\begin{figure}[h!]
\centering 
\includegraphics[width=0.9\linewidth]{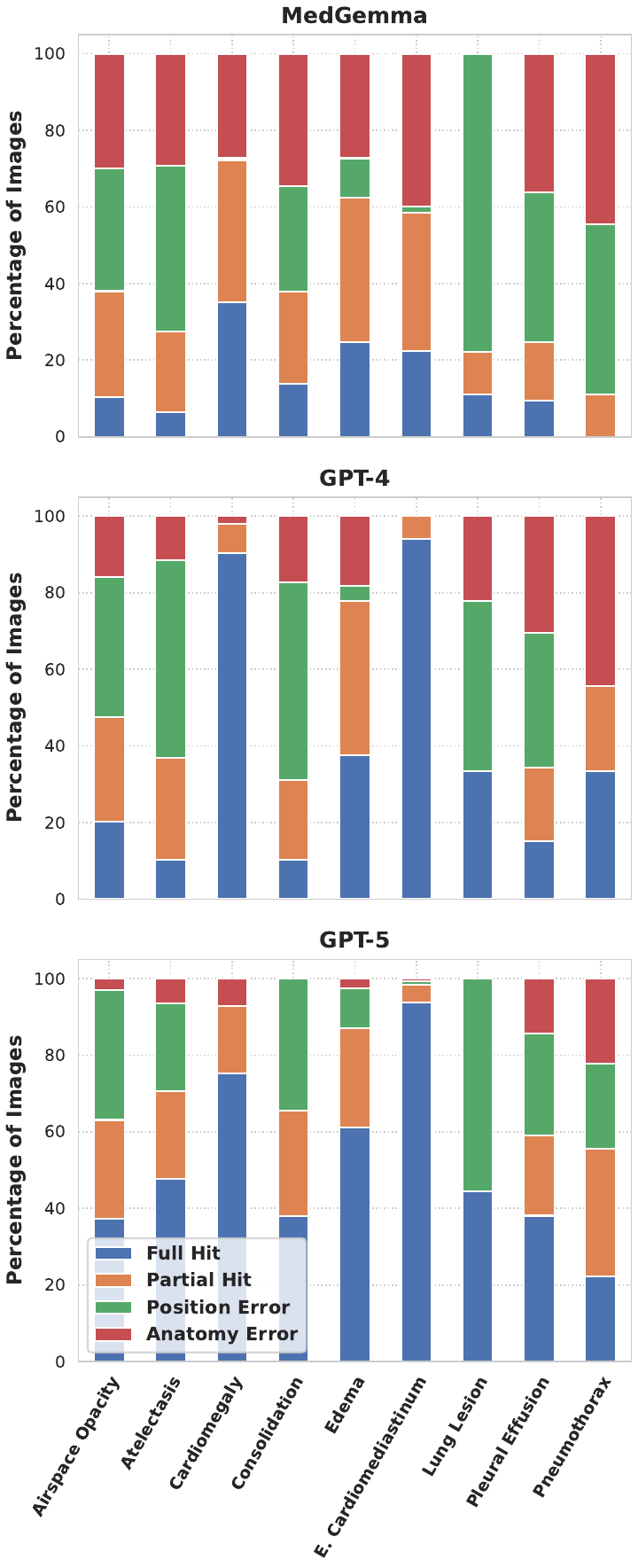}
\vspace{-10pt}
\caption{Error categorization. Each prediction on a frontal radiograph was categorized as a full hit ($\geq$50\% overlap), partial hit (0 $<$ overlap $<$ 50\%), position error (no overlap but plausible anatomy), and anatomy error (implausible anatomy).}
\label{fig:error_categorization} 
\end{figure}

To gain insights into the MLLMs' performance and general medical understanding, misses were reviewed and categorized as partial hits, position errors, or anatomy errors (see Methods for definitions).
The breakdown of the MLLM predictions by error category is displayed in Fig.~\ref{fig:error_categorization}.
For the two pathologies where GPT-4 and GPT-5 exhibited their highest performance (enlarged cardiomediastinum and cardiomegaly), the majority of misses were still partial hits for both models.
The distribution of error categories varied more widely for the other pathologies.
Despite a moderate average `full hit' rate, only 6.3\% of GPT-5's predictions represented an `anatomy error' on average across pathologies, indicating that the majority of misses were either a partial hit or at least corresponded to a plausible anatomical region. 
GPT-4 exhibited a higher rate of anatomy errors (18.0\% on average), with MedGemma exhibiting the most (29.9\% on average).
For both GPT-4 and GPT-5, the highest percentages of anatomy errors were observed for pleural effusion and pneumothorax, indicating that the models' predictions were relatively often localized to an incorrect anatomical structure (i.e., non-lung regions) for these pathologies.

\begin{figure*}[t]
\centering 
\includegraphics[width=1\textwidth]{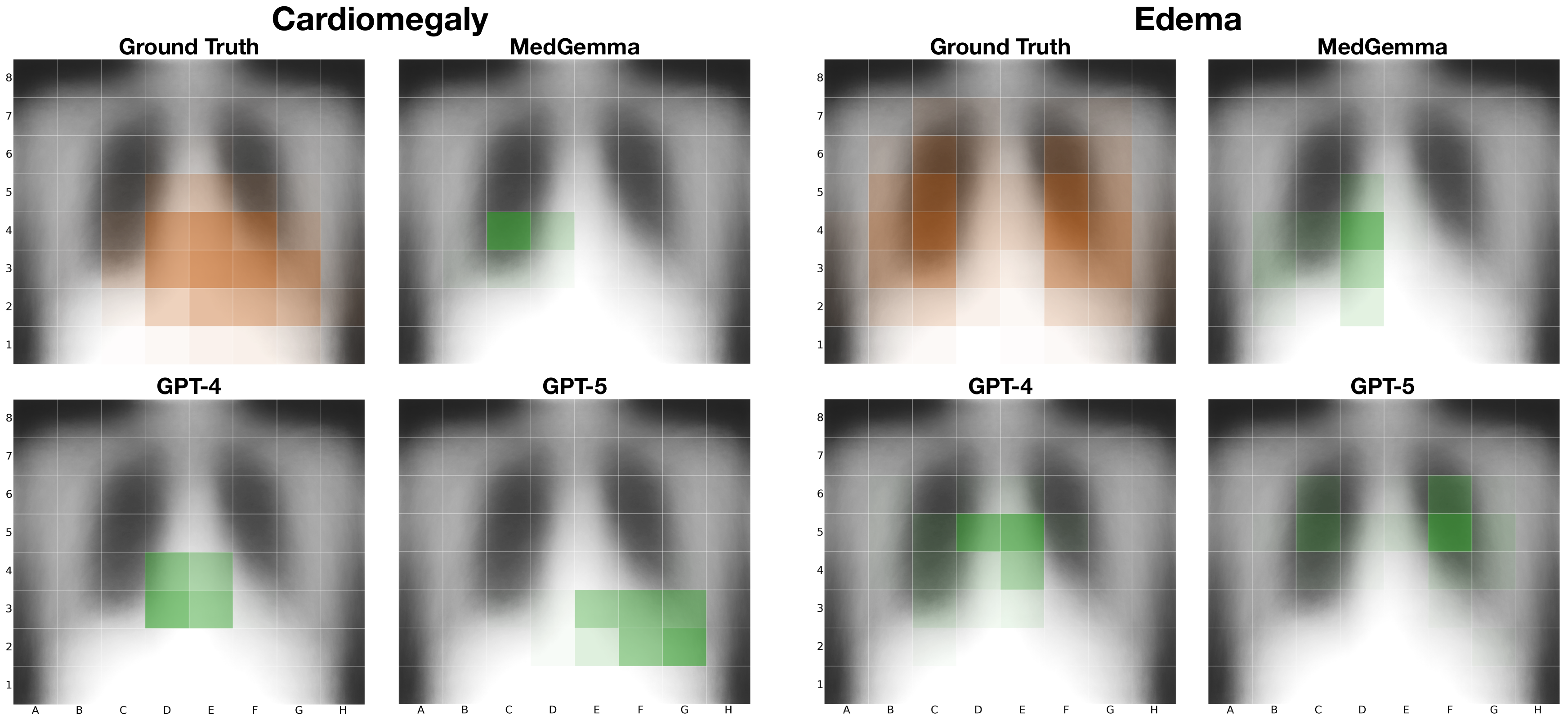} 
\caption{Example ground truth and prediction heatmaps. The heatmaps are computed over the frontal radiographs in the test set and are overlaid on the average image.}
\label{fig:example_heatmaps} 
\end{figure*}

\subsection{Visualization of Predictions}

Heatmaps of the models' predictions were generated to further understand the error patterns (Fig.~\ref{fig:example_heatmaps}, ~\ref{fig:appendix_heatmaps1},~\ref{fig:appendix_heatmaps2}). A separate heatmap was generated per pathology and model, summarizing the model's predictions for the given pathology across the dataset. 
To provide an aggregate anatomical reference, the heatmaps are visualized over the average image across the test set. Only frontal views were considered in generating the heatmaps. 

For cardiomegaly, GPT-4 consistently predicted central grid cells (Fig.~\ref{fig:example_heatmaps}). As cardiomegaly indicates an enlarged heart and the heart tends to appear towards the center of the image, the prediction heatmap aligns with the high performance of GPT-4 for this pathology. GPT-5 also predicted central grid cells for cardiomegaly, but demonstrated more variability that aligns with the ground truth distribution. 
Edema, indicating fluid in the lungs, varies more widely in its spatial position, though it often appears in the medial (towards center) regions of the lungs at the site of highest pulmonary vascular density.
For this pathology, GPT-4 consistently predicted a central grid cell, often overlaying the heart/mediastinum (Fig.~\ref{fig:example_heatmaps}), whereas GPT-5's predictions were more distributed over the lungs, explaining its higher performance on this pathology. 
In general, the heatmaps of GPT-5 more faithfully represented the ground truth pathology, explaining its relatively low percentage of anatomy errors (Fig.~\ref{fig:appendix_heatmaps1},~\ref{fig:appendix_heatmaps2}). 

\begin{figure*}[h]
\centering 
\includegraphics[width=.7\linewidth]{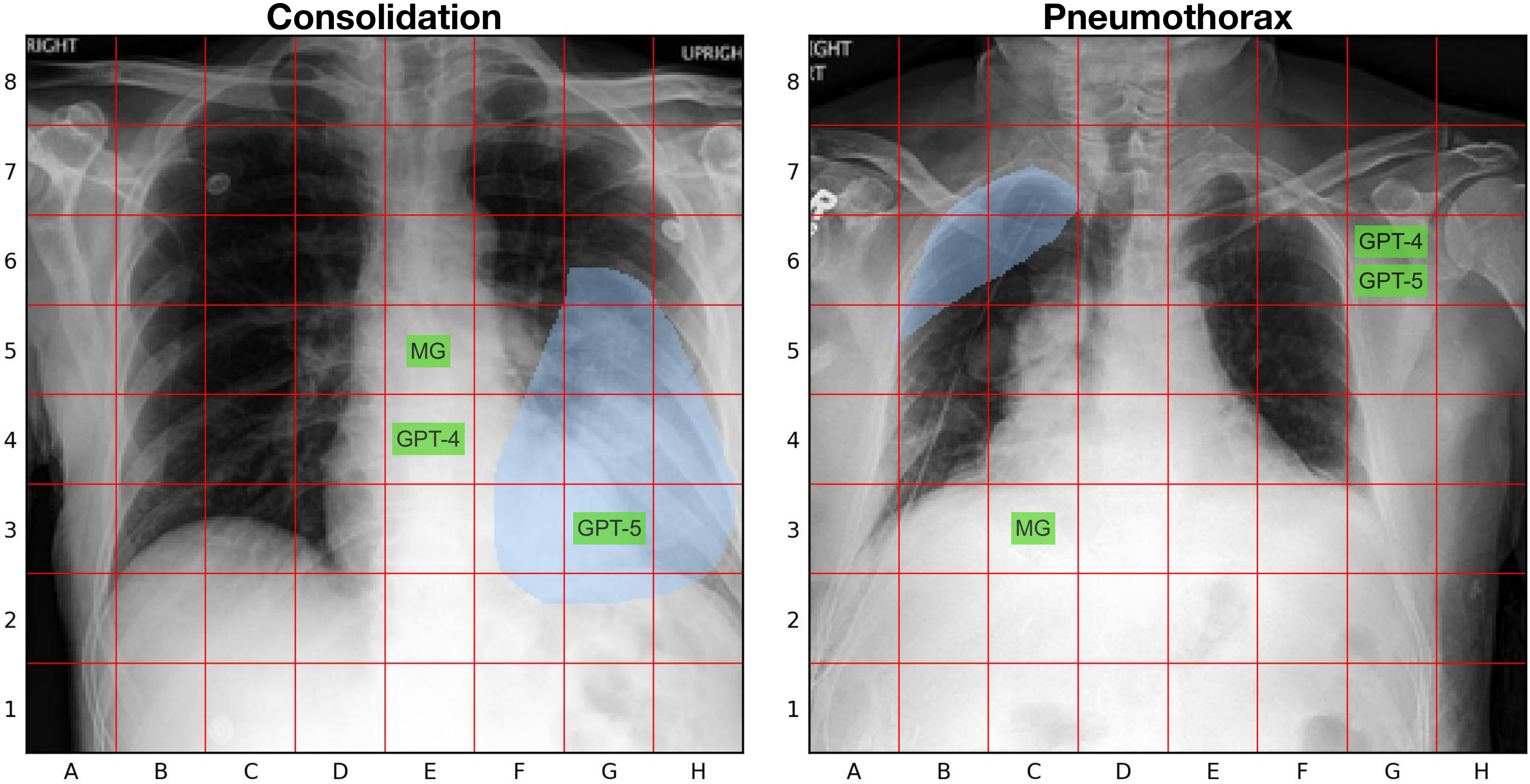} 
\caption{Example anatomy errors. a) A consolidation example where GPT-5 is correct but GPT-4's prediction overlays the heart/mediastinum. b) A pneumothorax example where both GPT-4 and GPT-5 predictions overlay the shoulder instead of the lungs.}
\label{fig:error_examples} 
\end{figure*}

Examining anatomy errors on an individual image basis revealed a spectrum of model misunderstanding. GPT-5 exhibited no anatomy errors for consolidation, which is a pathology that localizes to the lungs, but GPT-4 and MedGemma sometimes predicted grid cells overlaying the heart/mediastinum (Figure~\ref{fig:error_examples}). However, a dramatic error for GPT-5 is displayed in Figure~\ref{fig:error_examples}, wherein both GPT-5 and GPT-4 predicted the shoulder as the location of a pneumothorax (collapsed lung). 

\subsection{Prompt Variations}
Given the moderate performance, we explored whether different prompting strategies could lead to improvements. We considered two additional approaches: chain-of-thought (CoT) prompting wherein the models were instructed to think step-by-step, and few shot prompting wherein the models were provided three randomly selected examples for the given pathology. Additional details and the full prompts are contained in the Appendix.

The mean performance across pathologies for each prompting strategy is contained in Table \ref{mean_prompt_performance}, and the performance for each pathology contained in Appendix Table \ref{all_prompt_performance}. Overall, performance changes were modest. MedGemma exhibited the greatest relative improvement, from an average hit rate of 17.7\% to 22.5\% (27\% increase) and 32.0\% (81\% increase) for CoT and few shot prompting, respectively. On a per-pathology basis, few-shot prompting improved MedGemma's performance for all pathologies except lung lesion, with particularly large gains for atelectasis, consolidation, and pleural effusion ($>$130\% relative increases each). The improvement from CoT prompting was driven primarily by enlarged cardiomediastinum (191\% increase). Few shot prompting also elicited non-trivial predictions for MedGemma even when the temperature was set to 0, resulting in an average hit rate of 32.1\%, similar to that of the default temperature. Nonetheless, MedGemma's performances remained below the zero shot performances of GPT-4 and GPT-5. For GPT-4 and GPT-5, CoT prompting did not increase performance, but few shot prompting led to 4\% and 5\% relative increases, respectively. 
These increases were driven by edema and pleural effusion for GPT-4, and airspace opacity, cardiomegaly, and edema for GPT-5 (each with a $\ge$15\% relative hit rate improvement).

\begin{table}[h]
\centering
\caption{Mean hit rate across pathologies for each prompting strategy.}
\begin{tabular}{lcc}
\textbf{Model}            & \textbf{Prompt} & \textbf{\begin{tabular}[c]{@{}c@{}}Mean \\ Hit Rate\end{tabular}} \\ \hline
Random Baseline           & -               & 11.9                                                              \\
Human Benchmark           & -               & 80.1                                                              \\
CNN Baseline              & -               & 59.9                                                              \\ \hline
\multirow{3}{*}{MedGemma} & Zero Shot       & 17.7                                                              \\
                          & CoT             & 22.5                                                              \\
                          & Few Shot        & 32.0                                                              \\ \hline
\multirow{3}{*}{GPT-4}    & Zero Shot       & 39.1                                                              \\
                          & CoT             & 37.8                                                              \\
                          & Few Shot        & 40.8                                                              \\ \hline
\multirow{3}{*}{GPT-5}    & Zero Shot       & 49.7                                                              \\
                          & CoT             & 47.4                                                              \\
                          & Few Shot        & 52.1                                                             
\end{tabular}
\label{mean_prompt_performance}
\end{table}

\subsection{Comparing MedGemma to Gemma}
Although MedGemma is specifically designed for the medical domain, it is a relatively smaller model (27B parameters) compared to GPT-4 and GPT-5 (estimated to be at least 10x larger). To disentangle the effects of model size from domain-specific training, we evaluated the performance of Gemma 3 (27B), the base model on which MedGemma is built. Gemma performed similarly to MedGemma for zero shot prompting (average hit rate of 17.6\% vs. 17.7\%, respectively) and CoT prompting (25.0\% and 22.5\%); however, MedGemma's few shot performance was moderately higher (32.0\% vs. 25.5\%) with improvements observed for 7 out of 9 pathologies (Appendix Table \ref{all_prompt_performance}). Thus, MedGemma's low performance on the studied task may partly be attributable to its smaller size, where the domain-specific training appears to help with few shot prompting but not baseline performance. 

\section{Discussion} 
Beyond diagnosis, we evaluated the ability of frontier MLLMs to localize findings in chest radiographs.
We found that performance was mixed, with GPT-4 and GPT-5 performing relatively strong on pathologies that appear in consistent anatomical locations, but underperforming the CNN and human benchmark overall.
%Notably, these two pathologies tend to appear in consistent anatomical locations. %, while those where it struggled are more spatially variable. 
Nonetheless, GPT-5 showed strong improvements compared to GPT-4 on the more spatially variable pathologies, and its predictions were largely anatomically plausible even when it did not precisely localize the pathology.
%GPT-4 especially struggled with pathologies that are spatially variable and excelled at the two tend to appear in consistent anatomical locations
MedGemma, a medicine-specific model that includes chest radiographs in its training, exhibited the lowest performance overall and had a high proportion of anatomically-implausible predictions. Few shot prompting brought MedGemma's performance closer to the other models, with modest improvements for GPT-4 and GPT-5 as well. %The improvements with few shot prompting were especially observed for spatially variable pathologies.

Our findings have several actionable implications for clinical practice and AI development. First, while (M)LLMs have recently shown strong performance on question-answer challenges and diagnostic tasks \citep{Eriksen2023-ye, Suh2024-qw, nori2023capabilitiesgpt4medicalchallenge}, our results show that leading models struggle with fine-grained spatial reasoning. Smaller models such as MedGemma (and Gemma 3 27B on which it was built) may especially be limited in the ability to generalize to novel tasks, such as the one introduced here, compared to traditional benchmarks used in the field. Thus, the observed results may be partly attributable to the task itself rather than inherent medical knowledge, highlighting the importance of rigorous evaluation and adaptation strategies (e.g., few shot prompting) for specific use cases in clinical practice. While GPT-5's performance remained well below the human benchmark, the improvements in accuracy and medical understanding observed with GPT-5 compared to GPT-4 are encouraging. It remains to be seen whether these gains will continue to occur with future model versions, but these results at least suggest that large, general-purpose MLLMs are a promising approach compared to domain-specific vision-language models. In the meantime, agentic strategies that integrate the flexibility of LLMs with task-specific tools may be the best strategy, where recent work in the context of chest radiograph interpretation has demonstrated this possibility~\citep{medrax}. 
%Future work could evaluate how such agentic systems perform relative to GPT-4V in the task presented in this paper.

\paragraph{Limitations}
Our study has several limitations. Our analysis focused on one dataset given the need for localized annotations and meaningful benchmark comparisons. Future work could extend our analysis pipeline to other datasets and tasks. Furthermore, some pathologies in the CheXlocalize dataset have a small image count, leading to large error bars for the results for those pathologies. We included these pathologies because they still reveal meaningful model behavior, as illustrated in Fig. \ref{fig:error_examples} where GPT-4 and GPT-5 localize a lung pathology to the shoulder. Another challenge lies in the variability of ground truth annotation sizes across images and pathologies, which makes it difficult to define a hit rate metric that robustly captures all scenarios. Additionally, while interpretable, hit rate metrics sacrifice the flexibility and granularity offered by other localization metrics, such as IoU. Finally, although we explored different prompting strategies, more extensive prompt engineering could be explored in the future.

\paragraph{Conclusion} Systematically evaluating foundation models will be increasingly important as AI is integrated into clinical settings. 
Beyond our current findings, the proposed analysis pipeline can be leveraged to probe the spatial reasoning of foundation models more broadly.
A deeper understanding of the strengths and limitations of emerging models will be essential for guiding algorithm improvements and ensuring safe, effective clinical use.

\acks{W.L. acknowledges funding support from NIBIB award R21EB035247 and NLM award R01LM014775.}

\bibliography{main}

@ARTICLE{Eriksen2023-ye,
  title     = "Use of {GPT}-4 to diagnose complex clinical cases",
  author    = "Eriksen, Alexander V and Möller, Sören and Ryg, Jesper",
  journal   = "NEJM AI",
  publisher = "Massachusetts Medical Society",
  month     =  nov,
  year      =  2023,
  language  = "en"
}

@misc{openai2024gpt4technicalreport,
      title={GPT-4 Technical Report}, 
      author={{OpenAI}},
      year={2024},
      eprint={2303.08774},
      archivePrefix={arXiv},
      primaryClass={cs.CL},
      url={https://arxiv.org/abs/2303.08774}, 
}

@misc{nori2023capabilitiesgpt4medicalchallenge,
      title={Capabilities of GPT-4 on Medical Challenge Problems}, 
      author={Harsha Nori and Nicholas King and Scott Mayer McKinney and Dean Carignan and Eric Horvitz},
      year={2023},
      eprint={2303.13375},
      archivePrefix={arXiv},
      primaryClass={cs.CL},
      url={https://arxiv.org/abs/2303.13375}, 
}

@misc{gpt5,
  author       = {OpenAI},
  title        = {GPT-5},
  year         = {2025},
  howpublished = {\url{https://openai.com/gpt-5}}
}

@misc{medgemma,
  title={{MedGemma Technical Report}},
  author={Sellergren, Andrew and Kazemzadeh, Sahar and Jaroensri, Tiam and Kiraly, Atilla and Traverse, Madeleine and Kohlberger, Timo and Xu, Shawn and Jamil, Fayaz and Hughes, Cían and Lau, Charles and Chen, Justin and Mahvar, Fereshteh and Yatziv, Liron and Chen, Tiffany and Sterling, Bram and Baby, Stefanie Anna and Baby, Susanna Maria and Lai, Jeremy and Schmidgall, Samuel and Yang, Lu and Chen, Kejia and Bjornsson, Per and Reddy, Shashir and Brush, Ryan and Philbrick, Kenneth and Asiedu, Mercy and Mezerreg, Ines and Hu, Howard and Yang, Howard and Tiwari, Richa and Jansen, Sunny and Singh, Preeti and Liu, Yun and Azizi, Shekoofeh and Kamath, Aishwarya and Ferret, Johan and Pathak, Shreya and Vieillard, Nino and Merhej, Ramona and Perrin, Sarah and Matejovicova, Tatiana and Ramé, Alexandre and Riviere, Morgane and Rouillard, Louis and Mesnard, Thomas and Cideron, Geoffrey and Grill, Jean-Baptiste and Ramos, Sabela and Yvinec, Edouard and Casbon, Michelle and Buchatskaya, Elena and Alayrac, Jean-Baptiste and Lepikhin, Dmitry and Feinberg, Vlad and Borgeaud, Sebastian and Andreev, Alek and Hardin, Cassidy and Dadashi, Robert and Hussenot, Léonard and Joulin, Armand and Bachem, Olivier and Matias, Yossi and Chou, Katherine and Hassidim, Avinatan and Goel, Kavi and Farabet, Clement and Barral, Joelle and Warkentin, Tris and Shlens, Jonathon and Fleet, David and Cotruta, Victor and Sanseviero, Omar and Martins, Gus and Kirk, Phoebe and Rao, Anand and Shetty, Shravya and Steiner, David F. and Kirmizibayrak, Can and Pilgrim, Rory and Golden, Daniel and Yang, Lin},
  howpublished={arXiv preprint arXiv:2507.05201},
  year={2025},
  url={https://arxiv.org/abs/2507.05201}
}

@ARTICLE{McNamara2024-oc,
  title    = "The clinician-{AI} interface: intended use and explainability in
              {FDA}-cleared {AI} devices for medical image interpretation",
  author   = "McNamara, Stephanie L and Yi, Paul H and Lotter, William",
  journal  = "NPJ Digit Med",
  volume   =  7,
  number   =  1,
  pages    =  80,
  month    =  mar,
  year     =  2024,
  language = "en"
}

@ARTICLE{Zhou2024-qq,
  title     = "Evaluating {GPT}-{V4} ({GPT}-4 with vision) on detection of
               radiologic findings on chest radiographs",
  author    = "Zhou, Yiliang and Ong, Hanley and Kennedy, Patrick and Wu, Carol
               C and Kazam, Jacob and Hentel, Keith and Flanders, Adam and Shih,
               George and Peng, Yifan",
  journal   = "Radiology",
  publisher = "Radiological Society of North America (RSNA)",
  volume    =  311,
  number    =  2,
  pages     = "e233270",
  month     =  may,
  year      =  2024,
  language  = "en"
}

@ARTICLE{Saporta2022-lo,
  title     = "Benchmarking saliency methods for chest {X}-ray interpretation",
  author    = "Saporta, Adriel and Gui, Xiaotong and Agrawal, Ashwin and Pareek,
               Anuj and Truong, Steven Q H and Nguyen, Chanh D T and Ngo,
               Van-Doan and Seekins, Jayne and Blankenberg, Francis G and Ng,
               Andrew Y and Lungren, Matthew P and Rajpurkar, Pranav",
  journal   = "Nat. Mach. Intell.",
  publisher = "Springer Science and Business Media LLC",
  volume    =  4,
  number    =  10,
  pages     = "867--878",
  month     =  oct,
  year      =  2022,
  language  = "en"
}

@ARTICLE{Strotzer2024-wo,
  title     = "Toward foundation models in radiology? Quantitative assessment of
               {GPT}-{4V}'s multimodal and multianatomic region capabilities",
  author    = "Strotzer, Quirin D and Nieberle, Felix and Kupke, Laura S and
               Napodano, Gerardo and Muertz, Anna Katharina and Meiler, Stefanie
               and Einspieler, Ingo and Rennert, Janine and Strotzer, Michael
               and Wiesinger, Isabel and Wendl, Christina and Stroszczynski,
               Christian and Hamer, Okka W and Schicho, Andreas",
  journal   = "Radiology",
  publisher = "Radiological Society of North America",
  volume    =  313,
  number    =  2,
  pages     = "e240955",
  month     =  nov,
  year      =  2024,
  language  = "en"
}

@inproceedings{chexpert,
      title={CheXpert: A Large Chest Radiograph Dataset with Uncertainty Labels and Expert Comparison}, 
      author={Jeremy Irvin and Pranav Rajpurkar and Michael Ko and Yifan Yu and Silviana Ciurea-Ilcus and Chris Chute and Henrik Marklund and Behzad Haghgoo and Robyn Ball and Katie Shpanskaya and Jayne Seekins and David A. Mong and Safwan S. Halabi and Jesse K. Sandberg and Ricky Jones and David B. Larson and Curtis P. Langlotz and Bhavik N. Patel and Matthew P. Lungren and Andrew Y. Ng},
      year={2019},
        booktitle={AAAI}
}

@misc{medrax,
      title={MedRAX: Medical Reasoning Agent for Chest X-ray}, 
      author={Adibvafa Fallahpour and Jun Ma and Alif Munim and Hongwei Lyu and Bo Wang},
      year={2025},
      eprint={2502.02673},
      archivePrefix={arXiv},
      primaryClass={cs.LG},
      url={https://arxiv.org/abs/2502.02673}, 
}

@inproceedings{gradcam,
  author = {Selvaraju, Ramprasaath R. and Cogswell, Michael and Das, Abhishek and Vedantam, Ramakrishna and Parikh, Devi and Batra, Dhruv},
  booktitle = {ICCV},
  ee = {https://www.wikidata.org/entity/Q102362900},
  isbn = {978-1-5386-1032-9},
  pages = {618-626},
  publisher = {IEEE Computer Society},
  title = {Grad-CAM: Visual Explanations from Deep Networks via Gradient-Based Localization.},
  url = {http://dblp.uni-trier.de/db/conf/iccv/iccv2017.html#SelvarajuCDVPB17},
  year = 2017
}

@inproceedings{densenet,
  author = {Huang, Gao and Liu, Zhuang and van der Maaten, Laurens and Weinberger, Kilian Q.},
  booktitle = {CVPR},
  ee = {http://doi.ieeecomputersociety.org/10.1109/CVPR.2017.243},
  isbn = {978-1-5386-0457-1},
  pages = {2261-2269},
  publisher = {IEEE Computer Society},
  title = {Densely Connected Convolutional Networks},
  url = {http://dblp.uni-trier.de/db/conf/cvpr/cvpr2017.html#HuangLMW17},
  year = 2017
}

@ARTICLE{Brin2024-gp,
  title     = "Assessing {GPT}-4 multimodal performance in radiological image
               analysis",
  author    = "Brin, Dana and Sorin, Vera and Barash, Yiftach and Konen, Eli and
               Glicksberg, Benjamin S and Nadkarni, Girish N and Klang, Eyal",
  journal   = "Eur. Radiol.",
  publisher = "Springer Science and Business Media LLC",
  month     =  aug,
  year      =  2024,
  language  = "en"
}

@ARTICLE{Hayden2024-sc,
  title     = "Performance of {GPT}-4 with Vision on Text- and Image-based {ACR}
               Diagnostic Radiology In-Training Examination Questions",
  author    = "Hayden, Nolan and Gilbert, Spencer and Poisson, Laila M and
               Griffith, Brent and Klochko, Chad",
  journal   = "Radiology",
  publisher = "Radiological Society of North America",
  month     =  sep,
  year      =  2024,
  language  = "en"
}

@ARTICLE{Suh2024-qw,
  title    = "Comparing Diagnostic Accuracy of Radiologists versus {GPT}-{4V}
              and Gemini Pro Vision Using Image Inputs from Diagnosis Please
              Cases",
  author   = "Suh, Pae Sun and Shim, Woo Hyun and Suh, Chong Hyun and Heo, Hwon
              and Park, Chae Ri and Eom, Hye Joung and Park, Kye Jin and Choe,
              Jooae and Kim, Pyeong Hwa and Park, Hyo Jung and Ahn, Yura and
              Park, Ho Young and Choi, Yoonseok and Woo, Chang-Yun and Park,
              Hyungjun",
  journal  = "Radiology",
  volume   =  312,
  number   =  1,
  pages    = "e240273",
  month    =  jul,
  year     =  2024,
  language = "en"
}

@ARTICLE{Jin2024-ai,
  title    = "Hidden flaws behind expert-level accuracy of multimodal {GPT}-4
              vision in medicine",
  author   = "Jin, Qiao and Chen, Fangyuan and Zhou, Yiliang and Xu, Ziyang and
              Cheung, Justin M and Chen, Robert and Summers, Ronald M and
              Rousseau, Justin F and Ni, Peiyun and Landsman, Marc J and Baxter,
              Sally L and Al'Aref, Subhi J and Li, Yijia and Chen, Alexander and
              Brejt, Josef A and Chiang, Michael F and Peng, Yifan and Lu,
              Zhiyong",
  journal  = "NPJ Digit Med",
  volume   =  7,
  number   =  1,
  pages    =  190,
  month    =  jul,
  year     =  2024,
  language = "en"
}

\clearpage
\appendix

\section*{Appendix}\label{apd:first}

\subsection*{Prompts}
\subsubsection*{Baseline zero shot prompt}
\textbf{System:} You are an expert chest radiologist specializing in analyzing \{view\} chest X-rays. Your task is to precisely localize abnormalities using a grid overlay. \\
\textbf{User:} [Image] \\
This is a gridded \{view\} view of a chest X-ray. The abnormality `\{condition\}' is confirmed to be present in this image.
Your task:
\begin{enumerate}
    \vspace{-0.2cm}\item Identify the single grid cell where this abnormality - `\{condition\}' is the MOST prominent.
    \vspace{-0.2cm}\item Provide only the grid coordinate for this most representative cell. A grid coordinate is defined as a letter followed by a number. If the abnormality spans multiple cells, choose the cell that is most representative.
    \vspace{-0.2cm}\item Do not include any explanations or additional text in your response.
\end{enumerate}

\subsubsection*{Chain-of-Thought Prompt}
\textbf{System:} You are an expert chest radiologist specializing in analyzing {view} chest X-rays. Your task is to precisely localize abnormalities using a grid overlay. Approach this task methodically, using your expertise to analyze the image step-by-step. \\
\textbf{User:} [Image]\\
Your task:
\begin{enumerate}
\vspace{-0.2cm}\item Carefully examine the entire chest X-ray image, paying close attention to areas where `\{condition\}' typically manifests.
\vspace{-0.2cm}\item Identify all grid cells where you can observe signs of `\{condition\}'. List these cells in your thought process.
\vspace{-0.2cm}\item For each identified cell, briefly note the specific features or abnormalities you observe that are consistent with `\{condition\}'.
\vspace{-0.2cm}\item Compare the identified cells and their features. Consider which cell contains the most prominent or representative manifestation of `\{condition\}'.
\vspace{-0.2cm}\item If the condition spans multiple cells, determine which single cell encompasses the most significant part of the abnormality.
\vspace{-0.2cm}\item Based on your analysis, select the single grid coordinate (e.g., D5) that best represents the location of `\{condition\}' in this X-ray.
\vspace{-0.2cm}\item Provide your final answer as a single grid coordinate without any additional explanation or text.
\end{enumerate}
Example thought process:
``I observe signs of `\{condition\}' in cells C4, D4, and D5. In C4, there's [specific feature]. D4 shows [another feature]. D5 contains [feature]. Comparing these, D5 appears to be the most representative location for `\{condition\}'.
Final answer: D5''

\subsubsection*{Fewshot Prompt}
\textbf{System:} You are an expert chest radiologist specializing in analyzing chest X-rays. You are given a gridded \{view\} image of a chest X-ray with the abnormality `\{condition\}' confirmed to be present.\\
\textbf{User:} Zeroshot Prompt with [Example Image 1]\\
\textbf{Assistant:} Ground Truth grid coordinate for Example 1\\
\textbf{User:} Zeroshot Prompt with [Example Image 2]\\
\textbf{Assistant:} Ground Truth grid coordinate for Example 2\\
\textbf{User:} Zeroshot Prompt with [Example Image 3]\\
\textbf{Assistant:} Ground Truth grid coordinate for Example 3\\ \\
\textbf{User:} Zeroshot Prompt with [Test Image]

\vspace{0.2cm}
Few shot selection: Three random examples from the validation split with the same view position and pathology were selected. If there were not enough examples in the validation split, the remainder were selected randomly from the test split (excluding the image being evaluated).

%\newpage

%\clearpage

%\subsection{Performance Varying Grid Sizes}
\begin{figure*}[h!]
\centering 
\includegraphics[width=.75\textwidth]{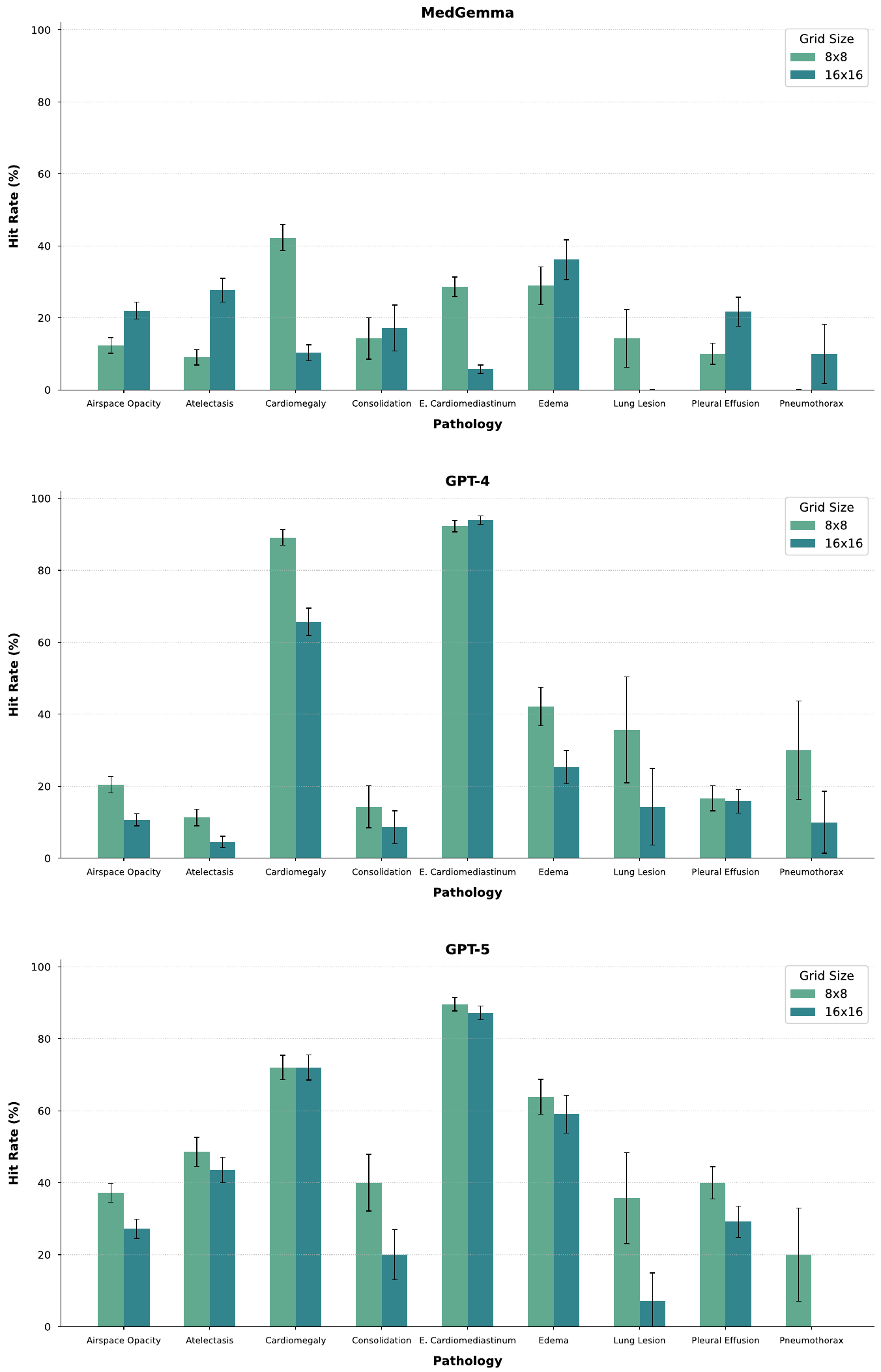} 
\caption{Performance by grid size. An 8x8 grid size was used for the main analysis. A higher resolution grid size (16x16) showed similar patterns between models and did not increase performance.}
\label{fig:gridsize_performance} 
\end{figure*}

%\subsection*{Prediction Heatmaps}
\begin{figure*}[h!]
\centering 
\includegraphics[width=\textwidth]{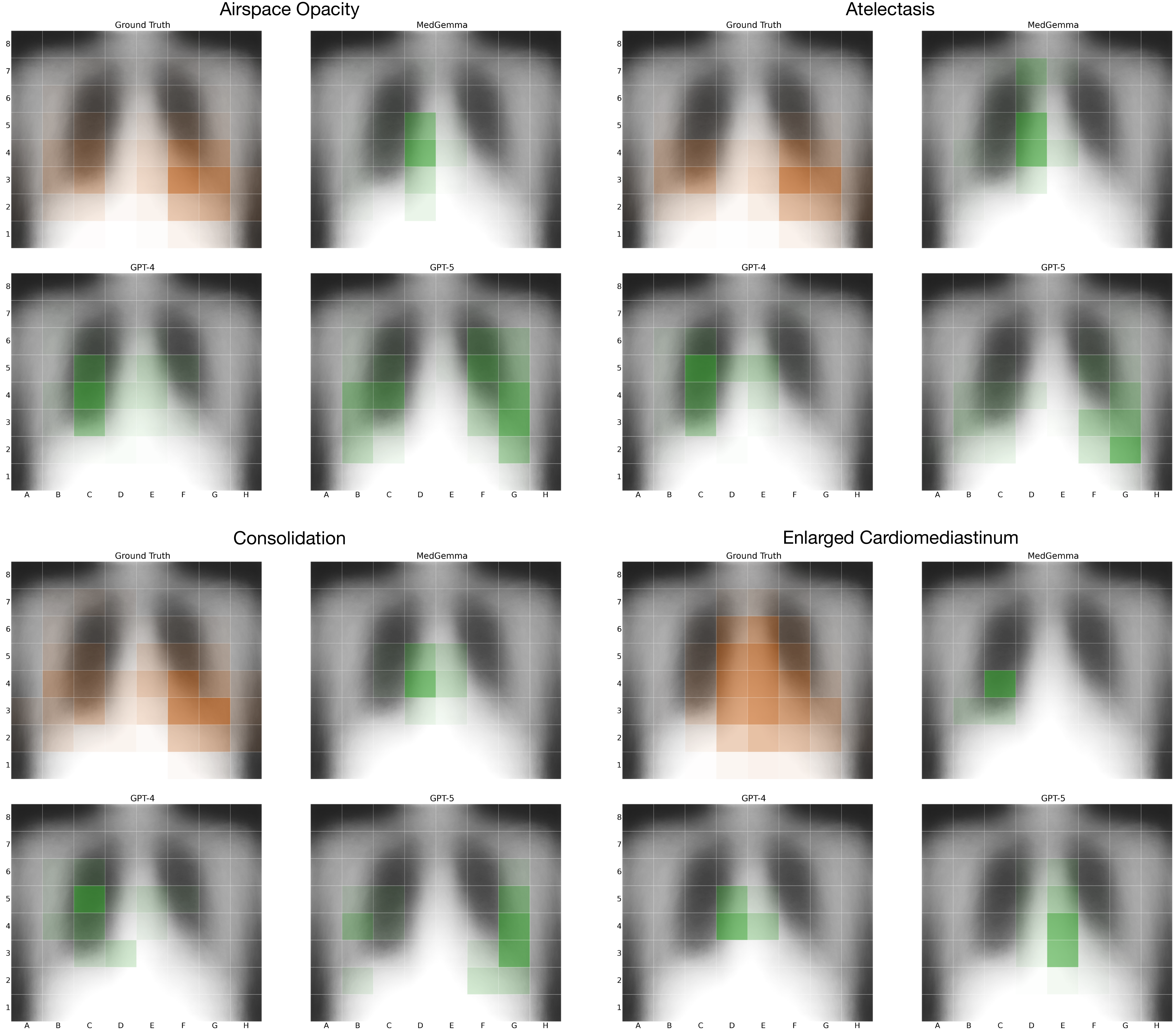} 
\caption{Ground truth and prediction heatmaps. The heatmaps are computed over the frontal radiographs in the test set and are overlaid on the average image. }
\label{fig:appendix_heatmaps1} 
\end{figure*}

\begin{figure*}[h!]
\centering 
\includegraphics[width=\textwidth]{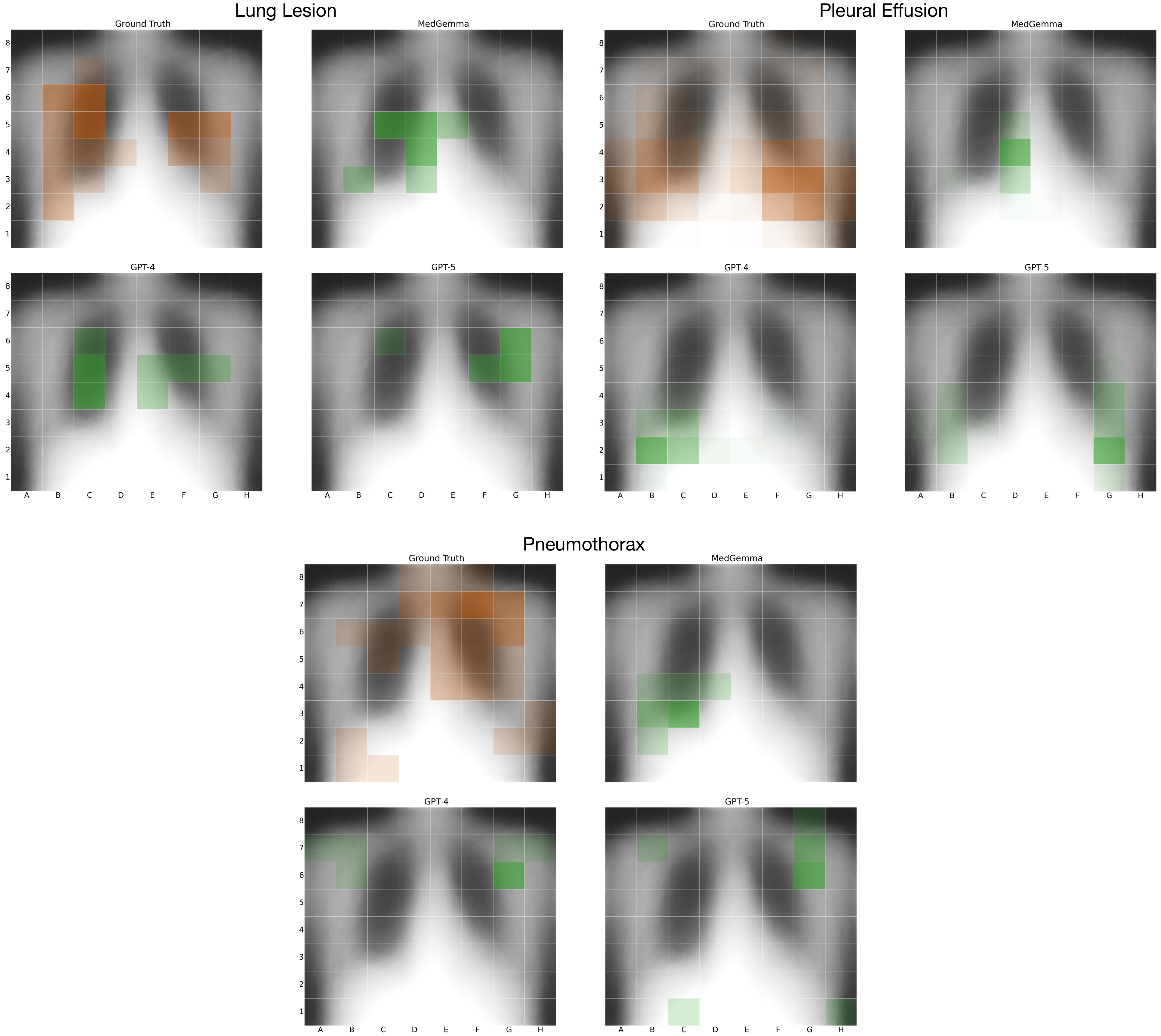} 
\caption{Ground truth and prediction heatmaps. The heatmaps are computed over the frontal radiographs in the test set and are overlaid on the average image. }
\label{fig:appendix_heatmaps2} 
\end{figure*}

\begin{figure*}[h]
\centering 
\includegraphics[width=0.8\textwidth]{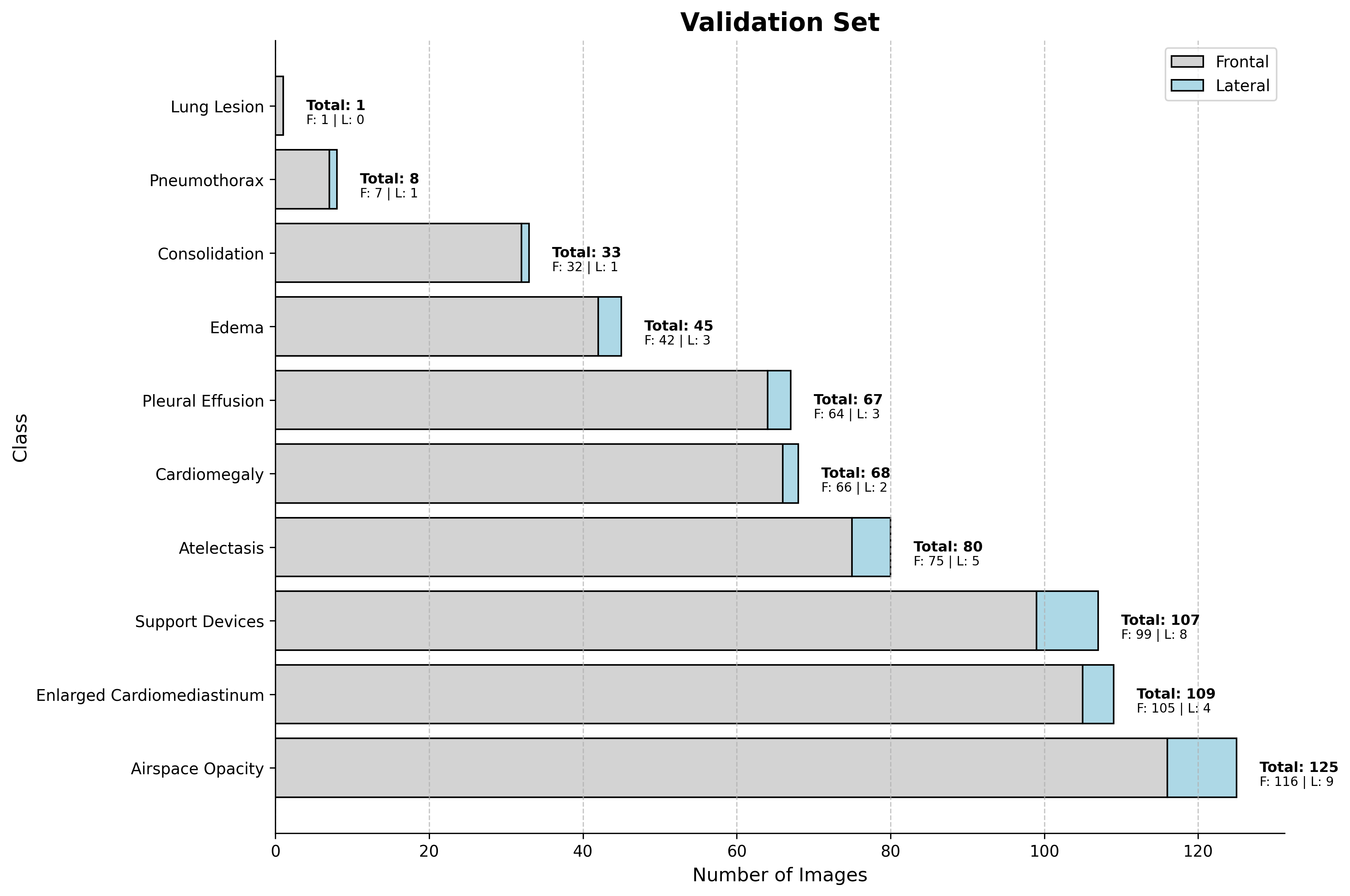} 
\caption{Image counts in CheXlocalize Validation Set.}
\label{fig:class_counts_val} 
\end{figure*}

\begin{figure*}[h]
\centering 
\includegraphics[width=0.8\textwidth]{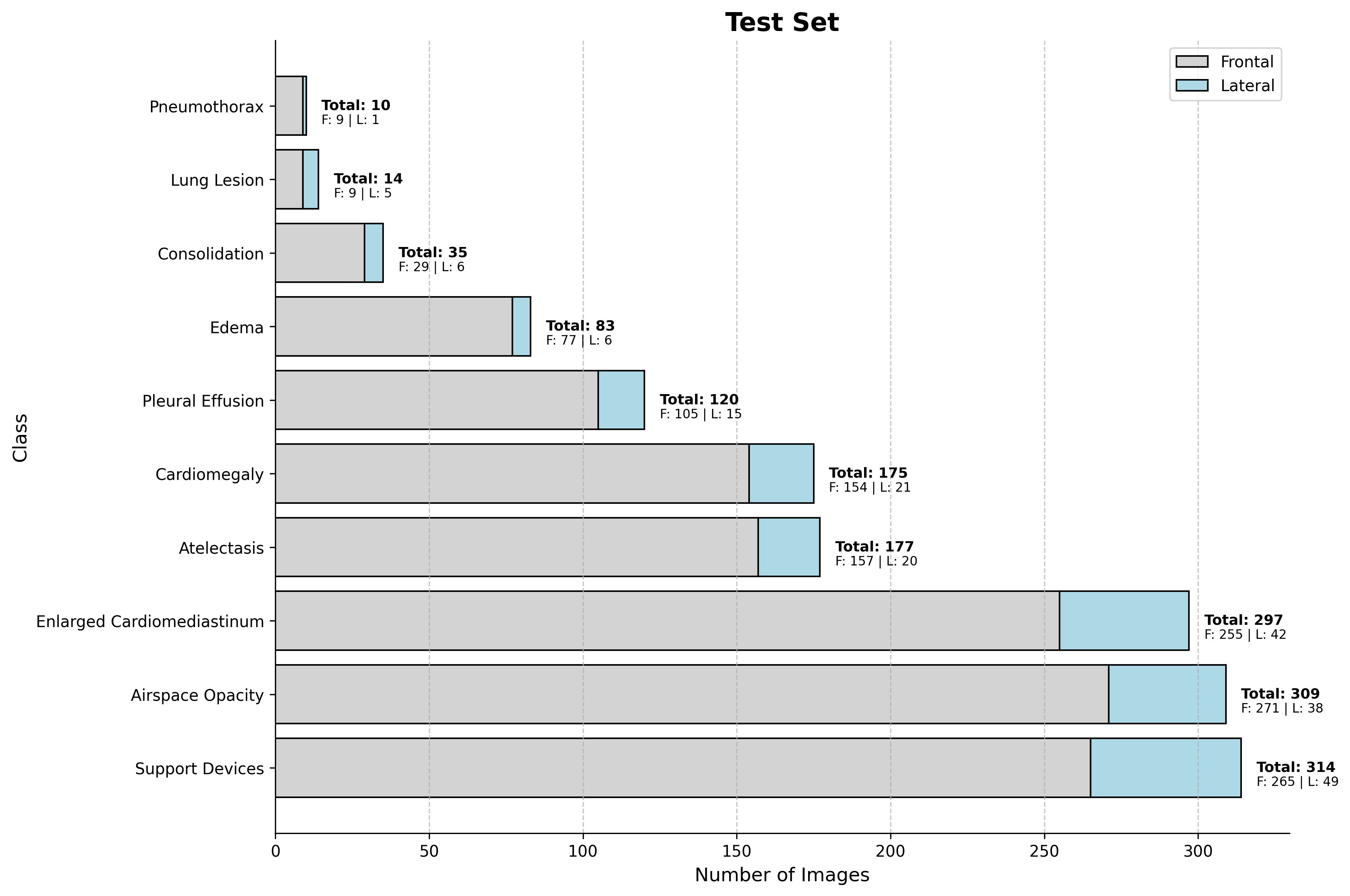} 
\caption{Image counts in CheXlocalize Test Set.}
\label{fig:class_counts_test} 
\end{figure*}

%\clearpage

\begin{sidewaystable*}[h!]
\small
\centering
\caption{Hit rate per pathology for each prompting strategy. Bold indicates the highest performance for the pathology; underline indicates the second highest.}
\begin{tabular}{lcccccccccc}
\textbf{Model}              & \textbf{Prompt} & \textbf{\begin{tabular}[c]{@{}c@{}}Airspace\\ Opacity\end{tabular}} & \textbf{Atelect.} & \textbf{Cardiomeg.} & \textbf{Consol.} & \textbf{Edema} & \textbf{E. Cardio.} & \textbf{\begin{tabular}[c]{@{}c@{}}Lung\\ Lesion\end{tabular}} & \textbf{\begin{tabular}[c]{@{}c@{}}Pleural\\ Effusion\end{tabular}} & \textbf{Pneumo.} \\ \hline
Random Baseline             & -               & 9.8                                                                 & 9.5               & 16.0                & 10.4             & 18.1           & 20.2                & 6.8                                                            & 8.6                                                                 & 7.5              \\
Human Benchmark             & -               & \textbf{55.9}                                                       & \textbf{87.0}     & \textbf{97.2}       & {\ul 51.0}       & \textbf{76.9}  & \textbf{95.7}       & \textbf{85.0}                                                  & \textbf{71.8}                                                          & \textbf{100.0}   \\
CNN Baseline                & -               & {\ul 49.8}                                                          & 50.1              & {\ul 90.3}          & \textbf{73.8}    & 74.6           & 81.8                & 29.0                                                           & {\ul 50.7}                                                       & {\ul 39.2}       \\ \hline
\multirow{3}{*}{MedGemma}   & Zero Shot       & 12.3                                                                & 9.0               & 42.3                & 14.3             & 28.9           & 28.6                & 14.3                                                           & 10.0                                                                & 0.0              \\
                            & CoT             & 6.1                                                                 & 7.9               & 37.7                & 17.1             & 31.3           & 83.2                & 0.0                                                            & 9.2                                                                 & 10.0             \\
                            & Few Shot        & 23.3                                                                & 23.7              & 57.7                & 37.1             & 43.4           & 62.6                & 7.1                                                            & 23.3                                                                & 10.0             \\ \hline
\multirow{3}{*}{GPT-4}      & Zero Shot       & 20.4                                                                & 11.3              & 89.1                & 14.3             & 42.2           & 92.3                & {\ul 35.7}                                                     & 16.7                                                                & 30.0             \\
                            & CoT             & 20.1                                                                & 9.0               & 81.1                & 22.9             & 45.8           & 90.2                & {\ul 35.7}                                                     & 25.0                                                                & 10.0             \\
                            & Few Shot        & 17.8                                                                & 11.3              & 85.7                & 20.0             & 54.2           & {\ul 92.9}          & {\ul 35.7}                                                     & 19.2                                                                & 30.0             \\ \hline
\multirow{3}{*}{GPT-5}      & Zero Shot       & 37.2                                                                & 48.6              & 72.0                & 40.0             & 63.9           & 89.6                & {\ul 35.7}                                                           & 40.0                                                                & 20.0             \\
                            & CoT             & 36.2                                                                & 44.6              & 72.6                & 31.4             & 61.4           & 91.9                & 21.4                                                           & 36.7                                                                & 30.0             \\
                            & Few Shot        & 44.7                                                                & {\ul 50.3}        & 84.6                & 45.7             & {\ul 74.7}     & 91.6                & 21.4                                                           & 45.8                                                                & 10.0             \\ \hline
\multirow{3}{*}{Gemma3 27B} & Zero Shot       & 12.0                                                                & 11.3              & 19.4                & 17.1             & 20.5           & 53.9                & 7.1                                                            & 7.5                                                                 & 10.0             \\
                            & CoT             & 6.1                                                                 & 9.6               & 33.1                & 17.1             & 50.6           & 83.5                & 21.4                                                           & 3.3                                                                 & 0.0              \\
                            & Few Shot        & 16.8                                                                & 21.5              & 42.9                & 5.7              & 34.9           & 73.1                & 14.3                                                           & 20.0                                                                & 0.0              \\ \hline
Image Count                 & -               & 309                                                                 & 177               & 175                 & 35               & 83             & 297                 & 14                                                             & 120                                                                 & 10              
\end{tabular}
\label{all_prompt_performance}
\end{sidewaystable*}

\end{document}